\documentclass[sigconf]{acmart}

\usepackage{subfig}
\usepackage{hyperref}
\usepackage{xspace}
\usepackage{tabularray}
\usepackage{multirow}
\usepackage{multicol}
\usepackage{bm}
\usepackage{tcolorbox}

\usepackage{makecell}
\usepackage{threeparttable} 

\usepackage[linesnumbered,algoruled,boxed,lined,ruled]{algorithm2e}
\SetKwInOut{Parameter}{Parameters}

\SetCommentSty{mycommfont}

\newcommand{\ourmethod}{MQE\xspace}
\AtBeginDocument{%
  \providecommand\BibTeX{{%
    \normalfont B\kern-0.5em{\scshape i\kern-0.25em b}\kern-0.8em\TeX}}}
    
\copyrightyear{2024} 
\acmYear{2024} 
\setcopyright{acmlicensed}
\acmConference[CIKM '24]{Proceedings of the 33rd ACM International Conference on Information and Knowledge Management}{October 21--25, 2024}{Boise, ID, USA} 
\acmBooktitle{Proceedings of the 33rd ACM International Conference on Information and Knowledge Management (CIKM '24), October 21--25, 2024, Boise, ID, USA} 
\acmDOI{10.1145/3627673.3679758} 
\acmISBN{979-8-4007-0436-9/24/10}
\begin{document}
\title{Noise-Resilient Unsupervised Graph Representation Learning \\ via Multi-Hop Feature Quality Estimation}

\author{Shiyuan Li}
\authornote{Both authors contributed equally to this research.}
\orcid{0000-0002-4381-7497}
\affiliation{
  \institution{Guangxi University}
  \city{Nanning}
  \country{China}
}
\email{shiy.li@alu.gxu.edu.cn}

\author{Yixin Liu}
\authornotemark[1]
\orcid{0000-0002-4309-5076}
\affiliation{
  \institution{Griffith University}
  \city{Gold Coast}
  \country{Australia}}
\email{yixin.liu@griffith.edu.au}

\author{Qingfeng Chen}
\orcid{0000-0002-5506-8913}
\authornote{Corresponding author.}
\affiliation{
  \institution{Guangxi University}
  \city{Nanning}
  \country{China}}
\email{qingfeng@gxu.edu.cn}

\author{Geoffrey I. Webb}
\orcid{0000-0001-9963-5169}
\affiliation{
  \institution{Monash University}
  \city{Melbourne}
  \country{Australia}}
\email{geoff.webb@monash.edu}

\author{Shirui Pan}
\orcid{0000-0003-0794-527X}
\affiliation{
 \institution{Griffith University}
 \city{Gold Coast}
 \country{Australia}}
\email{s.pan@griffith.edu.au}

\renewcommand{\shortauthors}{Li and Liu, et al.}

\begin{abstract}
  Unsupervised graph representation learning (UGRL) based on graph neural networks (GNNs), has received increasing attention owing to its efficacy in handling graph-structured data. 
However, existing UGRL methods ideally assume that the node features are noise-free, which makes them fail to distinguish between useful information and noise when applied to real data with noisy features, thus affecting the quality of learned representations.
This urges us to take node noisy features into account in real-world UGRL. With empirical analysis, we reveal that feature propagation, the essential operation in GNNs, acts as a ``double-edged sword'' in handling noisy features - it can both denoise and diffuse noise, leading to varying feature quality across nodes, even within the same node at different hops. 
Building on this insight, we propose a novel UGRL method based on \underline{M}ulti-hop feature \underline{Q}uality \underline{E}stimation (\ourmethod for short). Unlike most UGRL models that directly utilize propagation-based GNNs to generate representations, our approach aims to learn representations through estimating the quality of propagated features at different hops. 
Specifically, we introduce a Gaussian model that utilizes a learnable ``meta-representation'' as a condition to estimate the expectation and variance of multi-hop propagated features via neural networks. In this way, the ``meta representation'' captures the semantic and structural information underlying multiple propagated features but is naturally less susceptible to interference by noise, thereby serving as high-quality node representations beneficial for downstream tasks. 
Extensive experiments on multiple real-world datasets demonstrate that \ourmethod in learning reliable node representations in scenarios with diverse types of feature noise.

\end{abstract}



\maketitle

\section{Introduction}
Unsupervised graph representation learning (UGRL) aims to automatically extract low-dimensional vector representations from graph-structured data, eliminating the need for manual annotation~\cite{kipf2016variational,hamilton2017inductive}. 
The learned representations can be utilized for diverse downstream graph learning tasks, including node classification, node clustering, and graph classification~\cite{pan2018adversarially,zheng2024gnnevaluator,hou2022graphmae,liu2023reinforcement,zhang2023mutual,yang2022hierarchical}.  
Given their powerful capability to model graph data, graph neural networks (GNNs)~\cite{liang2024survey,pan2024integrating,koh2023psichic,zheng2024online} have become the de facto backbone models for various UGRL approaches~\cite{zhu2020deep,you2020graph,zhang2022costa,liu2022improved}. 
In recent years, UGRL has found applications across several domains, such as drug discovery, fraud detection, recommendation systems, and traffic forecasting~\cite{wu2020comprehensive,ge2020graph,wang2021multi,wang2024unifying}.

\begin{figure}[!t]
    \centering
    \includegraphics[scale=0.25]{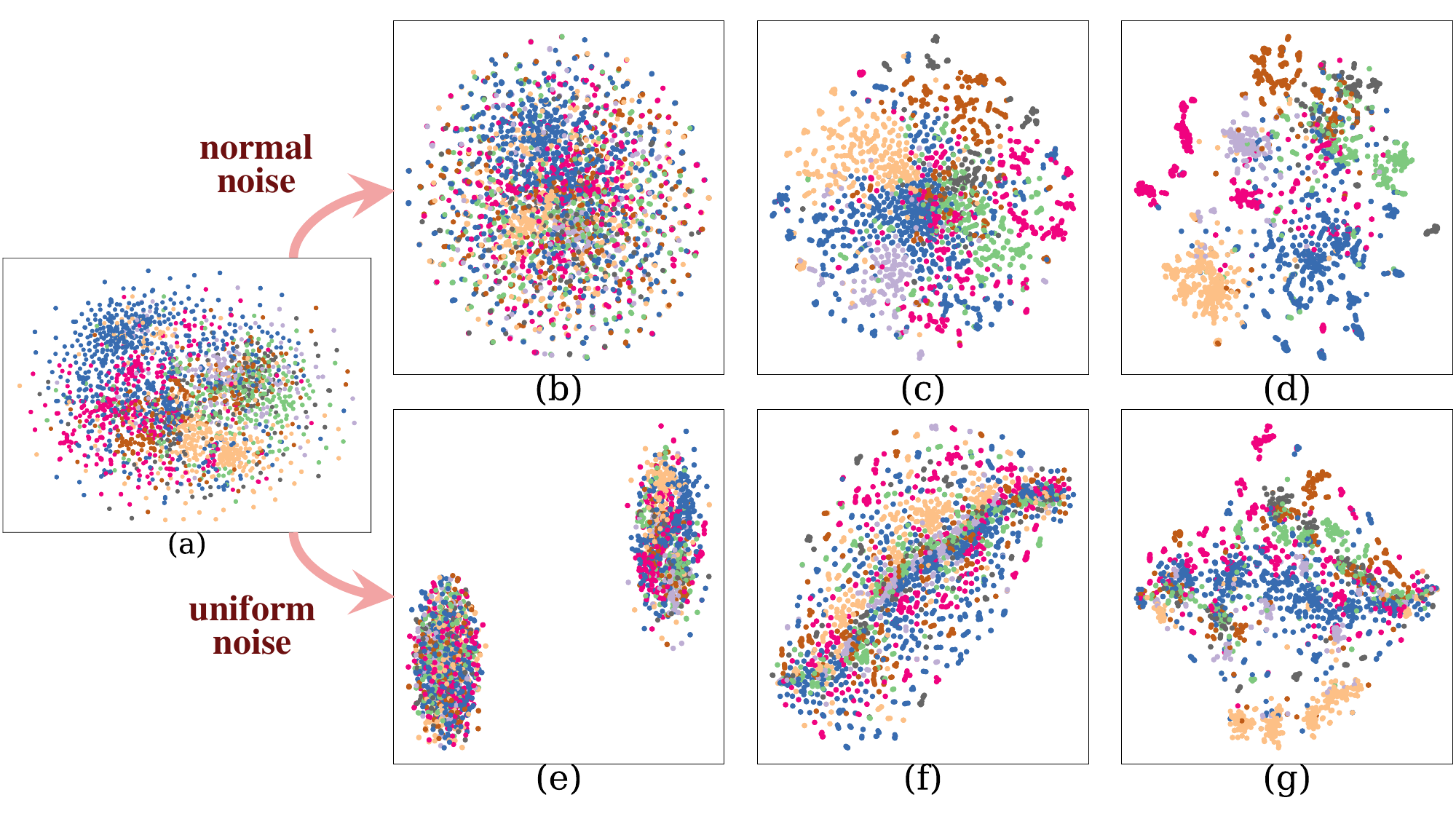}
    \vspace{-3mm}
    \caption{t-SNE~\cite{van2008visualizing} visualization of propagated features by symmetric normalized adjacency matrix of Cora dataset. (a) original features; (b)/(e) features after perturbation by noise; (c),(f)/(d),(g) noisy features after 2/16-step propagation.}
    \label{fig:tsne}
    \vspace{-5mm}
\end{figure}

The majority of UGRL models rely on the fundamental assumption that the observed node features are uncontaminated, thereby enabling the utilization of these features to furnish ample self-supervised signals for model training~\cite{zhu2021graph,hou2022graphmae}. Unfortunately, such an assumption is often invalid in real-world scenarios, as graph data extracted from complex systems frequently contains noisy, incomplete, and even erroneous features~\cite{zhou2023robust,li2024bi,li2012multiple,geng2021uncertainty,cai2024lgfgad,liu2024self}.
For example, considering privacy issues, users in many social networks may provide blank or even false information. Moreover, textual features may contain many spelling errors or informal expressions~\cite{zhang2023laennet,hu2013exploiting}. As illustrated in Fig.~\ref{fig:tsne} (a), (b) and (e), the presence of noise can significantly degrade both the quality and distribution of the original features, resulting in indistinguishable node features. As a result, these practical issues can lead to imperfect graph data characterized by noisy features, which poses a challenge to existing UGRL models. Specifically, contrastive methods often depend on prior-based augmentation to produce different views and construct positive and negative samples~\cite{zhu2020deep}. However, noisy features may obscure the boundaries between positive and negative samples, making it challenging for the model to accurately differentiate between them. Reconstruction-based methods, e.g., GraphMAE~\cite{hou2022graphmae}, usually involve the reconstruction of the original features, which can lead to model fitting to noisy information. Therefore, in the absence of reliable supervision signals, the noisy features inevitably prevent us from training an expressive UGRL model on real-world graph data. To remedy this shortcoming, a natural research question is raised: \textbf{\textit{How can we develop a robust UGRL approach that effectively addresses the presence of noisy features?}}

To answer this question, this paper first conducts a comprehensive analysis to investigate the performance of UGRL models when learning from noisy features. Through empirical discussions, we find that \textit{propagation, the fundamental operation for information aggregation in GNNs, is a double-edged sword in the case of noisy features}. On the one hand, under the assumption that noises are often high-frequency graph signals, propagation functions as a denoising mechanism in handling noisy graph data~\cite{nt2019revisiting}. From the spectral perspective, propagation operation is essentially equivalent to low-pass filtering in graph signal processing, which effectively suppresses the high-frequency portion of the message and thus reduces the effect of noise in signal passing. In this case, propagation is not only a way of information broadcasting, but also a signal processing filter that enhances the robustness of GNN-based UGRL models against the noisy feature issue. As illustrated in Fig.~\ref{fig:tsne} (c) and (d), propagation operations with different steps make noisy features more discernible, with features of the same category forming tightly knit clusters.

\begin{figure}[!t]
    \centering
    \includegraphics[width=0.75\columnwidth]{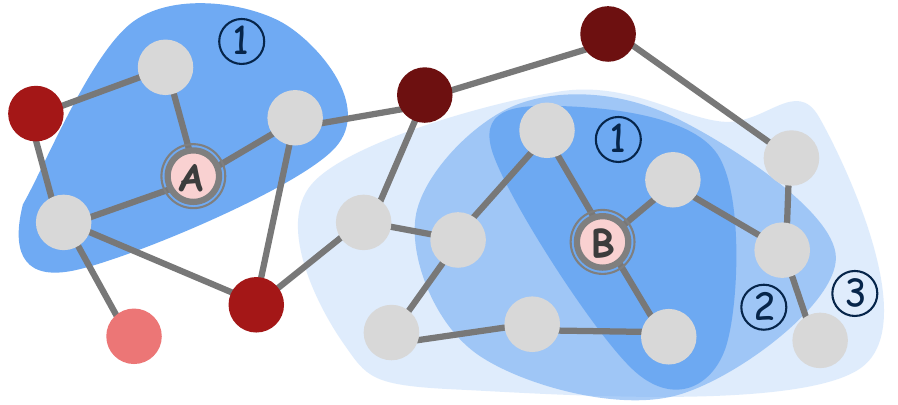}
    \vspace{-3mm}
    \caption{A toy example to illustrate the optimal propagation steps for different nodes. The color darkness of each node indicates its noise intensity (\textcolor[RGB]{160,160,160}{clean}-\textcolor[RGB]{221,124,121}{lightly noisy}-\textcolor[RGB]{100,25,21}{heavily noisy}). 
    For node A, a 1-step propagation that aggregates its 1-hop messages is beneficial for feature denoising. For node B, a larger number of propagation steps (e.g., 3) is preferred to generate a reliable representation by aggregating more neighboring information.}
    \label{fig:toy}
    \vspace{-5mm}
\end{figure}

On the other hand, propagation is not always the perfect solution for handling graphs with noisy features. As shown in Fig.~\ref{fig:tsne} (e), (f), and (g), for uniform noise, the propagation operation cannot remove the noise stably. In this case, the noisy signals will spread to the surrounding nodes along with the propagation process, affecting the representation quality of nodes with clean or slightly noisy features. Moreover, due to the uneven distribution of noise node positions and noise levels, the propagation steps to obtain high-quality representation for each node, which strike a balance between feature denoising and noise spreading, can vary significantly. Fig.~\ref{fig:toy} provides an example to illustrate this situation, where the representation quality at a fixed propagation step can be diverse across different nodes. More severely, in the scenario of UGRL, it is non-trivial to customize the best propagation step for each node without the guidance of ground-truth labels. 
Based on the above findings, an immediate core question is: \textit{\textbf{How to measure the representation quality obtained from different steps of propagation to derive higher-quality representations?}} 

To answer the question, we propose a robust and efficient UGRL model termed 
\underline{\textbf{M}}ulti-hop feature 
\underline{\textbf{Q}}uality
\underline{\textbf{E}}stimation (\ourmethod for short). Our theme is to adaptively learn the optimal representations of each node by explicitly estimating the quality of propagated features at different hops. More specifically, to exploit the inherited denoising capability of long-range propagation, we first perform multi-hop propagation on the original features as a prior observation. Then, we introduce a Gaussian model to estimate the quality of propagated features by modeling the truth signals and noise information under different hops. In the process of feature quality estimation, we introduce a learnable matrix $\mathbf{Z}$ termed ``meta representation'' to indicate the essential information of each node for different hops. The meta representation $\mathbf{Z}$ can serve as the high-quality node representations learned by \ourmethod, since it summarizes the most informative signals from propagated node features at different hops. Apart from learning reliable node representations, \ourmethod can also quantize the noise intensity of node features during quality estimation, which provides potential explanation and analysis for graph data. Notably, \ourmethod is different from the vast majority of UGRL approaches: rather than generating representations through propagation or GNNs, we attempt to estimate and fit different propagated features through learnable representations and neural networks. This characteristic enables \ourmethod to leverage the denoising merit multi-hop propagation while alleviating the side effect of noise spreading. Overall, the contributions of this paper are as follows:
\begin{itemize}
    \item \textbf{Problem.} We conduct comprehensive analyses to study the impact of noisy features on UGRL models, which provides an in-depth understanding of the research problem and model design.
    \item \textbf{Method.} We propose a novel method termed \ourmethod, which aims to estimate the quality of propagated features and adaptively learn high-quality representations from multi-hop information.     \item \textbf{Experiments.} We conduct extensive experiments on graph data with various types of feature noises. 
    The experimental results show that \ourmethod effectively learns reliable node representations and estimates the noise intensity of features.
\end{itemize}

\section{Notations \& Preliminary}
\subsection{Notation}
Given an attribute graph denoted as $\mathcal{G}=(\mathcal{V},\mathcal{E})$, where $\mathcal{V}=\{v_1, \cdots, v_n\}$ is the set of with $|\mathcal{V}|=n$ nodes and $\mathcal{E}$ is the set of edges. The adjacency matrix of the $\mathcal{G}$ is denoted as $\mathbf{A}$, where the $i,j$-th entry $a_{i j}=1$ if and only if the $i$-th and $j$-th nodes are connected (i.e. $(v_i,v_j) \in \mathcal{E}$), otherwise $\mathbf{a}_{i j}=0$. The symmetric normalization of the adjacency matrix is denoted by 
    $\hat{\mathbf{A}}=\widetilde{\mathbf{D}}^{-\frac{1}{2}} \tilde{\mathbf{A}} \widetilde{\mathbf{D}}^{-\frac{1}{2}}$, where $\tilde{\mathbf{A}}=\mathbf{A}+\mathbf{I}$ represents the adjacency matrix of the undirected graph with the addition of the self-loops and $\widetilde{\mathbf{D}}$ is the diagonal degree matrix of $\tilde{\mathbf{A}}$. We denote the neighbor set of node $v_i$ as $\mathcal{N}_i=\{v_j | (v_i,v_j) \in \mathcal{E}\}$. The attributes of nodes $\mathcal{V}$ can be represented by the node feature matrix $\mathbf{X} \in \mathbb{R}^{n \times d}$, where $d$ is the feature dimension and the $i$-th row $\mathbf{x}_i$ represents the feature vector of the $i$-{th} node $v_i$.

\subsection{Problem Definition}\label{sec:pdef}
Unsupervised graph representation learning (UGRL) aims to learn a mapping function $\mathcal{F}$ to obtain high-quality representations for all nodes in $\mathcal{G}$: $\mathcal{F}(\mathbf{A}, \mathbf{X}) \rightarrow \mathbf{Z} \in \mathbb{R}^{n \times f}$, where $\mathbf{Z}$ is the learned representation matrix, with each row $\mathbf{z}_i$ indicating the $f$-dimensional ($f \ll d$) representation vector of each node. These representations can be used for numerous downstream tasks, such as node classification and node clustering.

In mainstream UGRL settings, ones usually assume that the observed node features in the graph data are clean and informative enough to indicate the node contexts. However, in certain real-world scenarios, the feature vectors used for model training often contain noise, which can significantly impact the quality of learned representations and, consequently, the performance of downstream tasks.
To address this real-world challenge, our paper specifically focuses on the UGRL problem on graphs with noisy features. 

Formally, the graph feature matrix with noise can be defined by $\mathbf{X}=\widetilde{\mathbf{X}} + \mathbf{\Psi}$, where $\widetilde{\mathbf{X}}$ is the clean feature data (a.k.a. true signal in signal processing~\cite{nt2019revisiting}) and $\mathbf{\Psi}$ is the noise. Regarding the degree of noise within each node feature vector can be different, we further introduce \textit{noise intensity} to indicate the strength of noise of different nodes. To quantize noise intensity $s_i$ for node $v_i$, we can employ the modified L2 distance between true signal $\widetilde{\mathbf{x}}_i$ and observed signal $\mathbf{x}_i$, i.e.,  $s_i = \sqrt{ \frac{1}{d} \sum_{j=1}^{d} ({{x}_{ij}} - {\widetilde{{x}}_{ij}})^2}$. In our target scenario, a powerful UGRL model $\mathcal{F}(\mathbf{A}, \mathbf{X}) \rightarrow \mathbf{Z}$ is expected to learn high-quality representations $\mathbf{Z}$ from graph data with noisy features $\mathbf{X}$. Apart from representation learning, the proposed UGRL method, \ourmethod, can also estimate the noise intensity $s_i$ for each node.

\section{Design Motivation and Analysis}\label{MOTI}
\subsection{Can Propagation Alleviate Noisy Feature Problem in UGRL?}\label{sec3.1}

Propagation is a fundamental operation in GNNs for inter-node information communication along edges in graph-structured data. Since propagation has been proven to be effective in graph signal denoising on supervised GNNs~\cite{nt2019revisiting,liu2023learning,liu2024arc}, in the context of unsupervised scenarios, we are also curious about \textit{\textbf{whether propagation helps handle the noisy feature problem in UGRL}}.

\noindent\textbf{Theoretical Justification.} From the perspective of graph signal processing~\cite{ortega2018graph}, the mainstream GNNs (e.g., GCN~\cite{kipf2017semi} and SGC~\cite{wu2019simplifying}) can be regarded as low-pass filters for graph signals (i.e., node features). Specifically, Theorem 3 in~\cite{nt2019revisiting} indicates that by multiplying the graph signals with the propagation matrix, the low-frequency components within the graph signals are typically retained; Conversely, the high-frequency counterparts tend to be attenuated or smoothed out. Considering the basic assumption in signal processing that the true signals are usually low-frequency, the propagation operation in GNNs can inherently suppress the noise signals in features. Theoretically speaking, even in UGRL scenarios where supervised signals are absent, this principle still holds true.

\noindent\textbf{Intuitive Solution.} Given that propagation theoretically aids in feature denoising, a straightforward and effective approach to accentuate the denoising effect is to amplify the number of propagation steps~\cite{liu2023learning}. 
From the perspectives of both feature and structure, a larger propagation step can generally mitigate the noisy feature problem. 
On the feature side, a larger propagation step allows the model to consider a wider range of neighboring nodes, providing richer contextual knowledge for information aggregation while enabling noise filtering more effectively. On the structural side, such long-range propagation allows for communication between distant nodes, leveraging the existing graph structure knowledge more effectively in denoising. Preliminary visualization results in Fig.~\ref{fig:tsne} (b)-(d) verify that a larger propagation step can lead to more discriminative representation distribution on graph data with normal noise. However, Fig.~\ref{fig:tsne} (e)-(g) also illustrates that propagation is not always effective. Conversely, it may lead to noise diffusion thus affecting the quality of the representation of clean nodes.

\noindent\textbf{Empirical Analysis.} 
Based on the above discussion, we find that propagation may not always favor feature denoising. Therefore, to further quantify the impact of propagation more concretely, a natural question is: \textit{\textbf{ Would directly increasing the number of propagation steps in existing UGRL models suffice to learn informative representations from noisy features?}} To answer this question, with node classification as the downstream task, we assess the representation capability of prominent UGRL methods (specifically, DGI~\cite{velivckovic2018deep} and GRACE~\cite{zhu2020deep}) using both default (2), moderate (4) and large (16) propagation steps, across various noisy feature scenarios on the Cora dataset. To prevent the performance degradation~\cite{zhang2022model} of GNNs due to unnecessary transformation operations affecting the analysis results, SGC~\cite{wu2019simplifying}, a GNN with fix-step transformation, is uniformly employed as the backbone network for both DGI and GRACE. 

\begin{figure} [tbp]
	\centering
	\subfloat[Normal Noise\label{subfig:Cora_Normal_Degra}]{
		\includegraphics[scale=0.38]{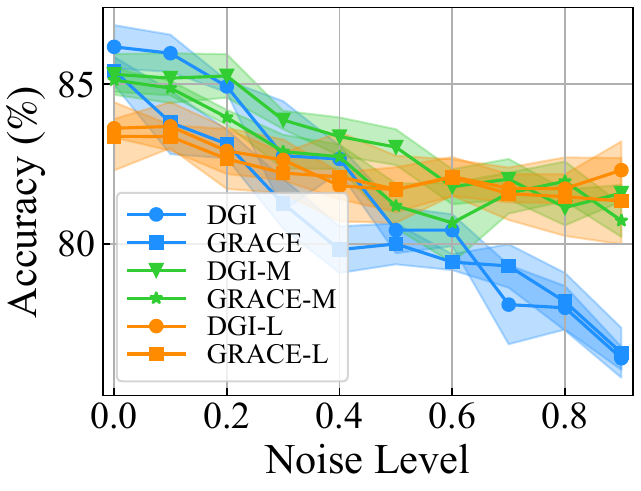}}\hfill
        \subfloat[Uniform Noise\label{subfig:Cora_Uniform_Degra}]{
		\includegraphics[scale=0.38]{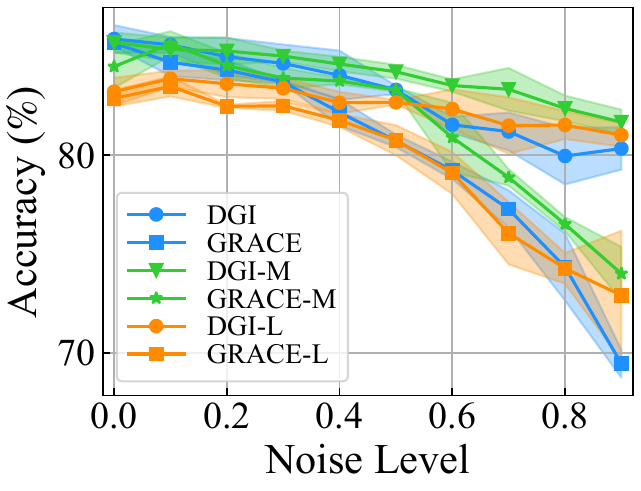}}\hfill
  \vspace{-2mm}
        \caption{The performance of UGRL models with default propagation step, moderate propagation (with suffix ``M'' and larger propagation step (with suffix
``L'') on the Cora dataset with different noise levels and types.}
\label{fig:DegraWNoise}
\vspace{-4mm}
\end{figure}

The experimental results are demonstrated in Fig.~\ref{fig:DegraWNoise}. 
We can summarize the following observations. 1) With the growth of noise level, the performance of DGI and GRACE significantly decreases. 2) For the scenarios with larger noise levels (i.e. $>0.5$), the models with larger propagation steps always perform better than their default counterparts. 3) For situations with clean or slightly noisy features, the gain from increasing the propagation step is small or even negative. 4) In uniform noise scenarios, increasing the propagation step only may further diffuse the noise, making the model performance degraded. 

\noindent\textbf{Discussion.} 
From the above empirical analysis, we can conclude that propagation is a double-edged sword - it can both denoise and diffuse noise. Specifically, increasing the propagation step is beneficial for UGRL models when the noise level is relatively high. Nevertheless, this is not a perfect solution, regarding the sub-optimal performance of large-step models in several scenarios. Therefore, a straightforward solution is to choose an optimal propagation step size to balance denoising and noise spreading. 

\subsection{Is there a Fixed Optimal Propagation Step?}\label{sec3.2}

In the above subsection, we delved into the effects of employing a larger propagation step to alleviate the issue of noisy features in UGRL. In this subsection, we discuss the impact of propagation from a more nuanced viewpoint: \textit{\textbf{Is there a fixed optimal propagation step for every node to acquire high-quality representation?}}

\noindent\textbf{Empirical Analysis.} 
To answer the aforementioned questions, we conducted an empirical experiment on Cora dataset with feature noise levels of 0.5. Specifically, we vary the propagation step of the SGC encoder in DGI and count the ``optimal propagation step'' for each node. Here, the term ``optimal propagation step'' refers to the specific propagation step that yields the highest average accuracy across multiple trials for the respective node. 

\begin{figure} [tbp]
	\centering
	\subfloat[Normal Noise\label{subfig:Cora_Normal_OptimalHop}]{
            \includegraphics[scale=0.38]{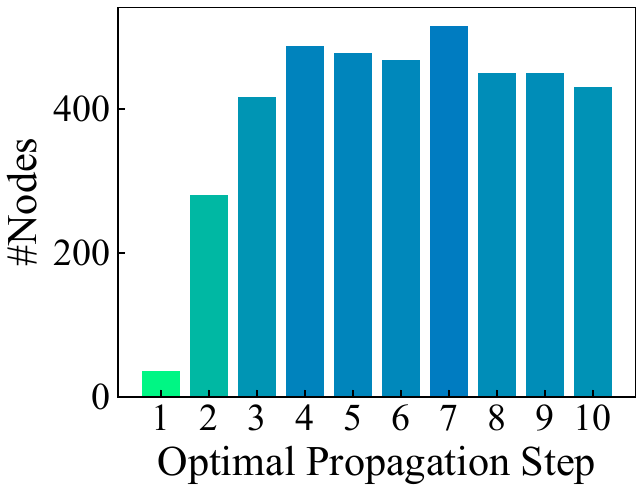}}\hfill
        \subfloat[Uniform Noise\label{subfig:Cora_Uniform_OptimalHop}]{
            \includegraphics[scale=0.38]{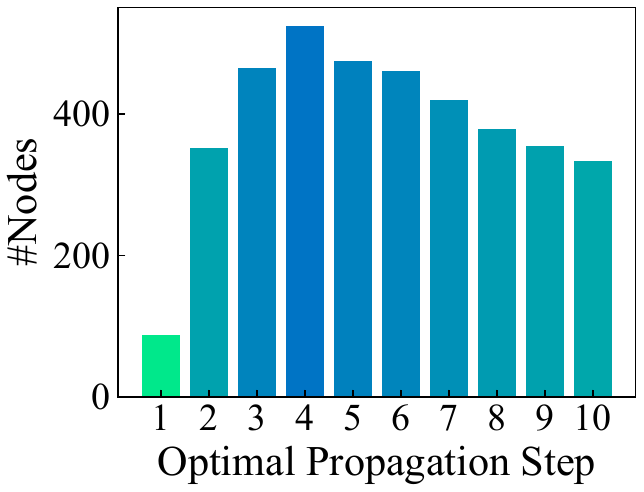}}\hfill
            \vspace{-2mm}
        \caption{The distribution of optimal propagation step for each node on Cora dataset with different noise types.}
        \vspace{-4mm}
	\label{fig:OptimalHop}
\end{figure}

The statistics of the distribution of optimal propagation steps are shown in Fig.~\ref{fig:OptimalHop}. Our observations are given as follows. 1)~The optimal propagation steps for different nodes can vary significantly. Specifically, for the majority of nodes, a larger propagation step usually yields better performance; however, there is still a small fraction of nodes that benefit from a smaller propagation step ($\leq 2$). 2)~Selecting a propagation step that universally benefits all nodes is difficult, as the distribution of optimal steps remains uniform. 3)~The distribution of optimal steps may vary depending on the type of feature noises. 

\noindent\textbf{Discussion.} 
With empirical verification, we identify the challenge of selecting an optimal propagation step that consistently enhances the representation quality for all nodes. \textit{\textbf{Why can larger propagation steps degrade the representation quality for several nodes?}} Regarding the presence of noisy features, we attribute such degradation to the spreading of noise through information propagation. Specifically, when a node's noisy features are difficult to denoise using low-pass filtering, it can greatly diminish the representation quality of its neighboring nodes. In this case, a node close to heavily noisy nodes may prefer a smaller propagation step to maintain the purity of its representations.

Furthermore, because nodes vary in their positions and feature noise intensity, the optimal propagation steps can vary across different nodes. This phenomenon is illustrated with a toy example in Fig.~\ref{fig:toy}. In this example, node B requires a larger propagation step to access more beneficial information, whereas node A may receive more noisy information when using a large propagation step. Such diversity, inherently, hinders us from using UGRL models with a singular propagation step to acquire high-quality representation for all nodes.

\begin{figure*} [tbp]
	\centering
            \includegraphics[scale=0.71]{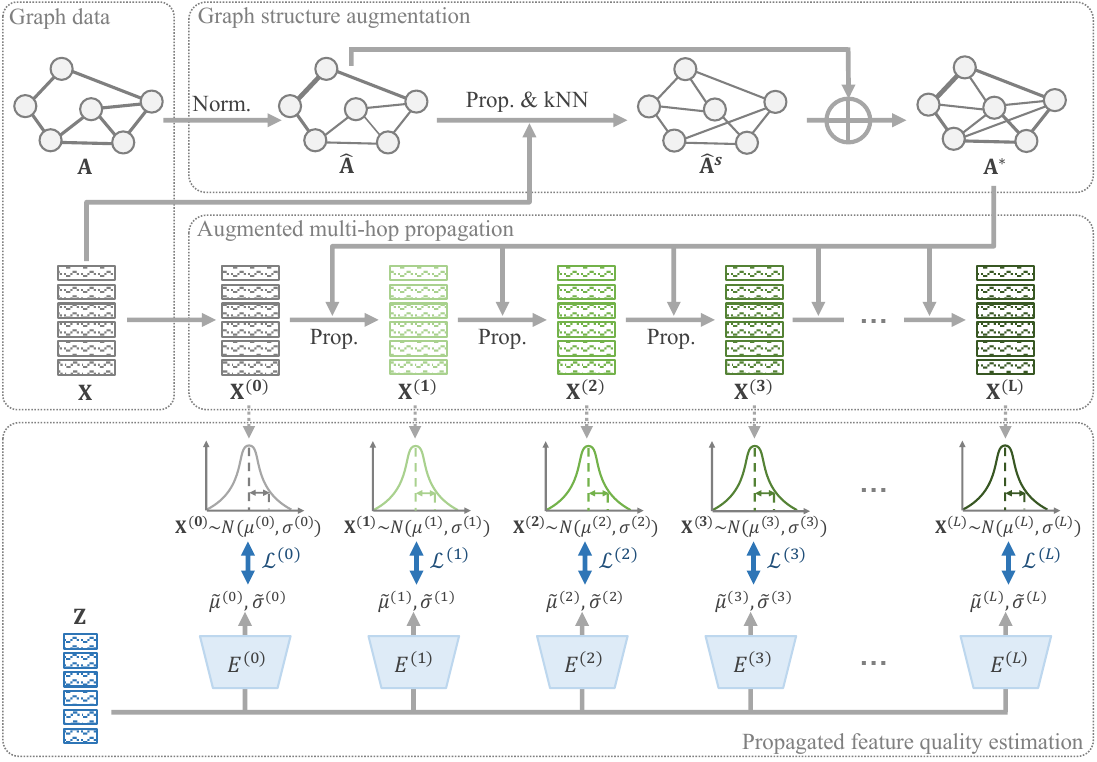}
            \vspace{-3mm}
        \caption{The pipeline of \ourmethod. First, a kNN-based \textit{graph structure augmentation} is conducted to generate the augmented adjacency matrix $\mathbf{A}^*$. Then, the propagated features are acquired by \textit{augmented multi-hop propagation} with $\mathbf{A}^*$. Afterward, in \textit{propagated feature quality estimation}, we take meta representations $\mathbf{Z}$ as the condition to estimate the mean $\mu$ and standard deviation $\sigma$ of the distribution of propagated features. $\mathbf{Z}$, finally, serves as the learned representations for downstream tasks.}
        \vspace{-4mm}
	\label{fig:Pipeline}
\end{figure*}

\noindent\textbf{Challenge.} 
The above deduction exposes that the quality of representation can undergo significant alterations based on the number of propagation steps utilized. In order to learn reliable representations for all nodes, a potential way is to fuse the representations acquired from multi-hop propagation~\cite{abu2019mixhop,zhu2020beyond,zhang2022graph}. 
Nevertheless, unlike the supervised models~\cite{abu2019mixhop,zhu2020beyond,zhang2022graph} that can learn to aggregate multi-hop information automatically, in UGRL models, it is non-trivial to decide the contributions of representations from different hops during aggregation without the guidance of labels. Moreover, multi-hop aggregation will also more or less allow noise information to mix into the final representation. In this case, a more promising solution is to explicitly estimate the quality of propagated features, and subsequently extract reliable representations from the high-quality ones.

\vspace{-1mm}
\begin{tcolorbox}[boxsep=0mm,left=2.5mm,right=2.5mm,colframe=black!40,colback=black!10]
\textbf{Summary:} With comprehensive analyses, we conclude that propagation serves as a double-edged sword in addressing the noisy feature problem in UGRL, embodying the trade-off between spectral graph signal denoising and spatial noise message spreading. Moreover, the diverse distribution of noisy nodes results in variations in the quality of node representations across different propagation steps. 
This diversity underscores the challenge of obtaining reliable representations through a fixed propagation step. In other words, the optimal steps vary among nodes rather than remaining fixed.
\end{tcolorbox}
\vspace{-3mm}

\section{Methodology}
According to the above analyses, the key to addressing the noisy feature problem in UGRL is to fully exploit the denoising mechanism of multi-hop propagation while mitigating the negative impact on representation quality caused by propagation. In line with this objective, we propose a novel UGRL method termed \underline{\textbf{M}}ulti-hop feature \underline{\textbf{Q}}uality \underline{\textbf{E}}stimation (\ourmethod for short) to tackle this challenging problem. Our core idea is to explicitly estimate the quality of propagated features at multiple propagation steps and to learn reliable node representations. 
The overall pipeline of \ourmethod is illustrated in Fig.~\ref{fig:Pipeline}. Specifically, we first conduct multi-hop feature propagation on an augmented graph structure, which enables us to acquire denoised features from different propagation steps. Then, we introduce Gaussian distribution to model the distribution of propagated features at multiple steps, enabling the estimation of their quality using the standard deviation $\sigma$. In \ourmethod, we utilize learnable latent variable $\mathbf{Z}$ called ``meta representations'' to encode the essential information of nodes. Afterward, neural network-based estimators $E$ are employed to approximate the expectation $\mu$ and standard deviation $\sigma$ of the propagated features based on $\mathbf{Z}$. Once \ourmethod is well-trained, the meta representations capture the true feature signals as well as the surrounding structural information of each node, and hence can serve as node representations for downstream tasks. The designs of \ourmethod will be introduced in the following subsections.

\subsection{Augmented Multi-Hop Propagation}

In Sec.~\ref{MOTI}, we recognized that long-distance propagation in UGRL can filter out high-frequency noisy information; however, relying solely on propagation-based GNNs to generate representations may pose a challenge due to the potential propagation of noise. To this end, in \ourmethod, we take the features acquired by multi-hop propagation as the target to be estimated rather than the learned representations. 

Specifically, the multi-hop propagated features can be calculated by iteratively. In each iteration, the propagated feature vector of the $i$-th node $v_i$ can be calculated by

\begin{equation}
    \mathbf{x}_i^{(\ell)} = \sum_{v_j\subset{\mathcal{N}_i}} \hat{a}_{i j} \mathbf{x}_j^{(\ell-1)},
\label{eq.4-1}
\end{equation}

\noindent where $\mathbf{x}_i^{(\ell)}$ is the propagated feature of node $v_i$ at the $\ell$-th propagation step, $\hat{a}_{i j}$ is the $i,j$-th entry of the symmetric normalization of the adjacency matrix $\hat{\mathbf{A}}_{i j}$, and $\mathcal{N}_i$ is the neighbor set of $v_i$. We define $\mathbf{x}_i^{(0)}=\mathbf{x}_i$ and calculate $\mathbf{x}_i^{(1)}, \cdots, \mathbf{x}_i^{(L)}$ via Eq.\eqref{eq.4-1} with a predefined maximum propagation step $L$.

It is worth noting that in \ourmethod, we employ a non-parameterized propagation scheme without any linear or non-linear feature transformations, which offers several advantages. Firstly, the propagated features can be precomputed during the preprocessing phase, as no learnable parameters are involved. As a result, the computational cost of model training can be acceptable, even if $L$ is large. Furthermore, the non-parameterized design avoids the potential risk of model degradation caused by redundant transformation operations, which have been shown to be detrimental in supervised GNNs~\cite{zhang2022model}.

\noindent\textbf{Graph Structure Augmentation.} 
Although propagating features on the original graph structure (defined by $\hat{\mathbf{A}}_{i j}$) is a default selection in the majority of GNNs~\cite{kipf2017semi,wu2019simplifying}, it may bring several issues in UGRL scenarios with noisy features. 
Given the uncertainty often present in real-world graph structures, the original graph inevitably includes noisy edges and misses key connections \cite{jin2020graph,chen2020iterative}. Therefore, directly executing propagation on $\hat{\mathbf{A}}_{i j}$ may result in ineffective feature denoising. Moreover, for the nodes that are isolated from other nodes in the original structure, propagation evidently fails to mitigate their noise characteristic issues. 

To address the aforementioned issues, in \ourmethod, we utilize graph structure augmentation~\cite{ding2022data,tan2023federated} to enhance the quality of the original structure, and then execute propagation on the augmented graph. Inspired by homophily assumption~\cite{kossinets2009origins}, we extend the neighbor set of each node by including new neighboring nodes with similar features. Specifically, we construct a k-nearest neighbor (kNN) graph $\mathbf{A}^s$ based on the propagated features for graph structure augmentation. Formally, we first add the propagated features at different steps together, and calculate their pair-wise cosine similarity by

\begin{equation}
    \mathbf{x}_i^* = \sum_{\ell = 0}^L \mathbf{x}_i^{(\ell)}, \quad
    \mathbf{a}_{i j}^s = \frac{\mathbf{x}_i^* \cdot \mathbf{x}_j^*}{||\mathbf{x}_i^*|| \cdot ||\mathbf{x}_j^*||},
\label{eq.4-2}
\end{equation}

\noindent where $\mathbf{a}_{i j}^s$ represents the similarity between the $i$-{th} and $j$-{th} nodes. Then, to preserve the sparsity of the augmented graph, we select the top-$k$ neighbors with the highest similarity as the extended neighbors of each node. Subsequently, symmetric normalization is applied to the sparsified graph, yielding the adjacency matrix of the supplementary graph, denoted by $\hat{\mathbf{A}}^s$. To harness both the information from the original graph and the supplementary graph collectively, we integrate them to create the augmented graph $\mathbf{A}^*$. Finally, we conduct multi-hop propagation on the augmented graph to obtain more robust propagated features:

\begin{equation}
    \mathbf{A}^* = \frac{\hat{\mathbf{A}} + \hat{\mathbf{A}}^s}{2}, \quad
    \hat{\mathbf{x}}_i^{(\ell)} = \sum_{v_j\subset{\mathcal{N}_i^*}} {a}_{i j}^* \hat{\mathbf{x}}_j^{(\ell-1)},
\label{eq.4-3}
\end{equation}

\noindent where ${a}_{i j}^*$ is the $i,j$-th entry of $\mathbf{A}^*$, ${\mathcal{N}_i^*}=\{v_j | {a}_{i j}^* > 0\}$ is the extended node neighbor set, and $ \hat{\mathbf{x}}_i^{(\ell)}$ is the propagated features for quality estimation in \ourmethod. Different from the stochastic graph augmentation in other UGRL methods~\cite{zhu2020deep,zhu2021graph}, the augmentation in \ourmethod is constant and non-parameterized. Hence, we only need to conduct the augmented multi-hop propagation once during the preprocessing phase.

\subsection{Propagated Feature Quality Estimation}\label{sec:pfqe}
With augmented multi-hop propagation, we can obtain propagated features containing the information of different receptive fields. However, as we discussed in Sec.~\ref{sec3.2}, the informativeness of the propagated features may vary across different propagation steps, resulting in their uncertain quality. To extract reliable representation from the propagated features at multiple steps, in \ourmethod, we introduce a Gaussian model-based estimation mechanism that aims to explicitly estimate the quality of propagated features. 

In the real world, sample features are usually obtained through a complex generative process, which can be approximated as a superposition of Gaussian distributions in large-scale data~\cite{choi2019gaussian}. Thus, to model the distribution of propagated features, we assume that the features at each step obey a Gaussian distribution, i.e., $\hat{\mathbf{x}}_i^{(\ell)} \sim N(\mu_i^{(\ell)}, (\sigma_i^{(\ell)})^2)$. Under such an assumption, the corresponding  probability density function can be written by

\begin{equation}
p\left(\hat{\mathbf{x}}_i^{\left(\ell\right)}\right)=\frac{1}{\sqrt{2 \pi\left(\sigma_i^{\left(\ell\right)}\right)^2}} e^{-\frac{\left(\hat{\mathbf{x}}_i^{\left(\ell\right)}-\mu_i^{\left(\ell\right)}\right)^2}{2\left(\sigma_i^{\left(\ell\right)}\right)^2}}. 
\label{eq.4-4}
\end{equation}

In Eq.~\eqref{eq.4-4}, the arithmetic mean $\mu_i^{(\ell)}$ indicates the expectation of true signals within the propagated feature vector $\hat{\mathbf{x}}_{i}^{(\ell)}$, while the standard deviation $\sigma_i^{(\ell)}$ reflects the quality of $\hat{\mathbf{x}}_{i}^{(\ell)}$. More concretely, a larger $\sigma_i^{(\ell)}$ indicates a lower quality of $\hat{\mathbf{x}}_{i}^{(\ell)}$, as it suggests that the features likely experience a greater shift from the true signal.

Given the Gaussian distribution-based modeling for the propagated features, we can transfer the quality estimation task to a variational prediction task for the distribution of $\hat{\mathbf{x}}_i^{(\ell)} \sim N(\mu_i^{(\ell)}, (\sigma_i^{(\ell)})^2)$. Based on the above modeling, our goal is to encode fruitful information of each node from multiple propagation steps into a unified node representation. Thus, for each node $v_i$ we define a learnable vector known as the ``meta-representation'' $\mathbf{z}_i$ as a consistent condition shared across multiple propagation steps. In the feature quality estimation phase, the meta-representation $\mathbf{z}_i$ can be viewed as a compact latent variable encapsulating node attributes and contextual information. 

Specifically, in \ourmethod, $\mathbf{z}_i$ is initialized with random values and then updated concurrently with the model parameters. Given the meta representation $\mathbf{z}_i$ as condition, we establish two estimator networks $E^{(\ell)}_\mu(\cdot)$ and $E^{(\ell)}_\sigma(\cdot)$ to estimate $\mu_i^{(\ell)}$ and $\sigma_i^{(\ell)}$, respectively. In practice, $E^{(\ell)}_\mu(\cdot)$ and $E^{(\ell)}_\sigma(\cdot)$ are both 2-layer multilayer perceptron (MLP) networks. Formally, the conditional likelihood for the $\ell$-step propagated features can be written by

\begin{equation}\label{eq.4-5}
p\left(\hat{\mathbf{x}}_i^{(\ell)} \mid \mathbf{z}_i\right)=N\left(\tilde{\mu}_i^{(\ell)},\left(\tilde{\sigma}_i^{(\ell)}\right)^2\right)=\frac{1}{\sqrt{2 \pi\left(\tilde{\sigma}_i^{(\ell)}\right)^2}} e^{-\frac{\left(\hat{\mathbf{x}}_i-\tilde{\mu}_i^{(\ell)}\right)^2}{2\left(\tilde{\sigma}_i^{(\ell)}\right)^2}},
\end{equation}

\noindent where $\tilde{\mu}_i^{(\ell)}=E_\mu^{(\ell)}(\mathbf{z_i})$ and $\tilde{\sigma}_i^{(\ell)}=E_\sigma^{(\ell)}(\mathbf{z_i})$ are the estimated mean and standard deviation acquired from the estimators, respectively.

To implement effective feature quality estimation, we optimize the estimators $E = \{E^{(0)}, \cdots, E^{(L)}\} = \{E^{(0)}_\mu, E^{(0)}_\sigma,  \cdots, E^{(L)}_\mu, E^{(L)}_\sigma\}$ for every propagation steps as well as the meta representations $\mathbf{Z} = [\mathbf{z}_1^\intercal,\cdots,\mathbf{z}_n^\intercal]$ with a variational reconstruction loss. Specifically, we employ a negative logarithmic reconstruction likelihood loss function, which can be written by

\begin{equation}\label{eq.4-7}
\begin{aligned}
\mathcal{L} & = \sum_{\ell=1}^L \mathcal{L}^{(\ell)} =  \sum_{\ell=1}^L \sum_{i=1}^n-\ln \left(p\left(\mathbf{x}_i^{(\ell)} \mid \mathbf{z}_i\right)\right) \\
& =\sum_{\ell=1}^L \sum_{i=1}^n \frac{\left(\mathbf{x}_i^{(\ell)}-\tilde{\mu}_i^{(\ell)}\right)^2}{2\left(\tilde{\sigma}_i^{(\ell)}\right)^2}+\ln \left(\tilde{\sigma}_i^{(\ell)}\right),
\end{aligned}
\end{equation}

\noindent where a constant term in the original expression is ignored since it does not affect the optimization of the model. In the loss function, the first term aims to minimize the reconstruction error between $\tilde{\mu}_i^{(\ell)}$ and the propagated features $\mathbf{x}_i^{(\ell)}$, with a tolerance controlled by $\tilde{\sigma}_i^{(\ell)}$. Concretely, a larger value of $\tilde{\sigma}_i^{(\ell)}$ denotes a more relaxed constraint on the reconstruction error, consequently implying a diminished quality of $\mathbf{x}_i^{(\ell)}$. Our experiments detailed in Sec.~\ref{sec:qual_exp} affirm that $\tilde{\sigma}_i^{(0)}$ effectively serves as an indicator for the noise intensity of features. The second term serves as a regularization component aimed at averting collapse to a trivial solution, where $\sigma$ becomes excessively large for all propagated features. 

\noindent\textbf{$\mathbf{Z}$ as Final Representations.} 
During the model training process, $\mathbf{Z}$ progressively and adaptively amalgamates information from various propagated features, ultimately evolving into concise and informative representations for each node. 
As an accurate estimation of multi-hop propagated features, $\mathbf{Z}$ encompasses not only the true feature signals but also the knowledge from neighboring nodes.
Hence, during the inference phase, we can directly utilize $\mathbf{Z}$ as the learned representations by \ourmethod, employing them in various downstream tasks.

\noindent\textbf{Complexity Analysis.} The time complexity of \ourmethod primarily consists of two main components: pre-processing and model training. For the pre-processing phase which involves feature propagation and kNN-based augmentation, the complexity is $\mathcal{O}\left(Lmd+n^2d\right)$ ($m=|E|$ is the number of edges) in total, where the first term is for propagation and the second term is for kNN computation. For each training iteration, complexity mainly arises from the two-layer MLP-based estimators. Specifically, the complexity of each $E_\mu^{(\ell)}$ and $E_\sigma^{(\ell)}$ are $\mathcal{O}\left(nh(f+d)\right)$ and $\mathcal{O}\left(nhf\right)$, respectively, where $h$ and $f$ are the dimensions of hidden layer and meta representation. Therefore, the overall training time complexity is $\mathcal{O}\left(Lnh(f+d)\right)$ in total.

\section{Experiments}

\begin{table*}
\centering
\caption{Node classification accuracy. The results of the winner and runner-up are bolded and underlined, respectively.}
\vspace{-3mm}
\resizebox{1\textwidth}{!}{
\begin{tabular}{p{1.7cm}|p{1.95cm}|p{1.8cm}<{\centering}|p{1.8cm}<{\centering} p{1.8cm}<{\centering} p{1.8cm}<{\centering}|p{1.8cm}<{\centering} p{1.8cm}<{\centering} p{1.8cm}<{\centering}}  
\toprule
\multirow{2}{*}{Dataset} & \multirow{2}{*}{Method} & \multirow{2}{*}{Clean} & \multicolumn{3}{c|}{Normal Noise ($\alpha=0.5$)} & \multicolumn{3}{c}{Uniform Noise ($\alpha=0.5$)} \\ 
\cline{4-9}
& & & \multicolumn{1}{c}{$\beta=0.3$} & \multicolumn{1}{c}{$\beta=0.5$} & $\beta=0.8$ & \multicolumn{1}{c}{$\beta=0.3$} & \multicolumn{1}{c}{$\beta=0.5$} & $\beta=0.8$ \\ 
\midrule
\multirow{9}{*}{Cora}
& GAE & 81.21±0.87 & 72.24±1.09 & 70.17±1.50 & 63.09±1.30 & 79.28±1.01 & 77.94±1.13 & 71.46±2.42 \\
& VGAE & 77.05±1.07 & 70.14±0.87 & 68.11±0.34 & 66.53±0.79 & 76.17±1.15 & 73.11±0.93 & 64.47±2.51 \\ \cline{2-9}
& GRACE & \underline{85.86±0.63} & 82.82±0.61 & 81.02±0.53 & 79.04±2.54 & 82.74±0.65 & 81.24±0.66 & 79.46±0.62 \\
& DGI & 85.38±0.76 & 81.85±0.79 & 79.44±0.56 & 75.28±1.56 & 84.29±0.29 & 82.75±0.91 & 81.07±1.16 \\
& MVGRL & 85.63±0.75 & \underline{83.93±0.71} & 80.96±0.53 & 79.13±0.71 & \underline{85.35±0.65} & \underline{84.10±0.74} & \underline{82.54±0.83} \\
& COSTA & 85.66±0.56 & 82.74±0.95 & \underline{81.62±0.81} & 79.76±1.09 & 84.35±0.82 & 83.42±0.93 & 81.60±0.88 \\ \cline{2-9}
& GraphMAE & 85.61±0.52 & 82.79±1.21 & 81.08±1.41 & \underline{79.84±0.68} & 84.56±0.62 & 83.88±0.71 & 81.81±0.36 \\
& MaskGAE & 84.25±0.80 & 77.48±1.10 & 72.59±0.31 & 67.80±0.48 & 82.65±0.64 & 81.71±0.89 & 78.75±0.53 \\ \cline{2-9}
& \textbf{\ourmethod(Ours)} & \textbf{85.97±0.42} & \textbf{84.71±0.64} & \textbf{83.27±0.52} & \textbf{82.16±0.55} & \textbf{85.44±0.51} & \textbf{84.89±0.54} & \textbf{83.76±0.28} \\ 
\midrule
\multirow{9}{*}{CiteSeer}
& GAE & 68.76±0.95 & 56.09±1.55 & 53.25±0.68 & 48.45±0.96 & 67.36±0.85 & 63.37±1.97 & 40.74±7.73 \\
& VGAE & 63.70±0.84 & 48.91±1.64 & 42.58±2.24 & 37.34±1.21 & 62.04±0.86 & 57.20±2.89 & 46.60±2.76 \\ \cline{2-9}
& GRACE & 73.21±0.71 & 69.08±0.85 & 66.94±1.18 & 64.94±0.63 & 72.15±0.45 & 70.32±0.73 & 65.55±0.72 \\
& DGI & 73.39±0.92 & 65.07±0.67 & 60.50±1.57 & 55.82±0.93 & \underline{72.50±0.78} & \underline{71.32±0.60} & 68.10±0.76 \\
& MVGRL & 73.19±0.51 & 67.62±1.35 & 64.36±0.42 & 62.14±0.70 & 72.04±0.58 & 70.98±0.44 & 66.88±1.39 \\
& COSTA & 72.16±0.59 & 68.98±0.40 & 58.08±1.66 & 53.23±1.48 & 69.41±0.96 & 68.78±1.24 & 62.69±0.70 \\ \cline{2-9}
& GraphMAE & \underline{73.82±0.48} & \underline{70.15±0.24} & \underline{68.90±0.26} & \underline{65.90±0.47} & 70.38±0.61 & 68.97±0.52 & \underline{68.50±0.23} \\
& MaskGAE & 72.73±0.45 & 64.68±0.82 & 58.71±0.88 & 53.67±0.63 & 71.86±0.56 & 69.91±0.93 & 65.68±0.52 \\ \cline{2-9}
& \textbf{\ourmethod(Ours)} & \textbf{74.64±0.53} & \textbf{71.53±0.88} & \textbf{69.44±0.98} & \textbf{66.37±1.27} & \textbf{73.79±0.73} & \textbf{72.64±1.00} & \textbf{72.25±0.58} \\ 
\midrule
\multirow{6}{*}{PubMed}
& GRACE & 86.10±0.28 & 72.34±0.52 & 66.28±0.57 & 60.31±0.32 & 81.12±0.38 & 75.95±0.42 & 70.11±0.79 \\
& DGI & 87.08±0.17 & 77.19±0.21 & 71.77±0.44 & 66.21±0.33 & \underline{84.49±0.33} & 81.77±0.30 & 78.30±0.28 \\
& COSTA & 86.98±0.33 & 75.45±0.45 & 69.47±0.35 & 64.86±0.37 & 83.37±0.34 & 79.80±0.37 & 74.99±0.27 \\ \cline{2-9}
& GraphMAE & 85.94±0.16 & \underline{80.49±0.25} & \underline{77.42±0.24} & \underline{76.99±0.31} & 84.23±0.22 & \underline{82.63±0.37} & \underline{82.01±0.23} \\
& MaskGAE & \underline{87.48±0.20} & 73.89±0.30 & 67.81±0.38 & 62.44±0.57 & 84.11±0.39 & 80.30±0.36 & 75.72±0.58 \\ \cline{2-9}
& \textbf{\ourmethod(Ours)} & \textbf{87.73±0.19} & \textbf{82.71±0.30} & \textbf{81.15±0.23} & \textbf{80.37±0.38} & \textbf{85.43±0.23} & \textbf{84.23±0.22} & \textbf{82.52±0.23} \\ 
\bottomrule
\toprule
\multirow{2}{*}{Dataset} & \multirow{2}{*}{Method} & \multirow{2}{*}{Clean} & \multicolumn{3}{c|}{Normal Noise ($\alpha=0.5$)} & \multicolumn{3}{c}{Uniform Noise ($\alpha=0.5$)} \\ 
\cline{4-9}
& & & \multicolumn{1}{c}{$\beta=1$} & \multicolumn{1}{c}{$\beta=5$} & $\beta=10$ & \multicolumn{1}{c}{$\beta=1$} & \multicolumn{1}{c}{$\beta=5$} & $\beta=10$ \\ 
\midrule
\multirow{6}{*}{Photo}
& GRACE & 92.62±0.23 & 91.65±0.16 & 88.30±0.36 & 87.50±0.67 & 90.96±0.51 & 87.94±0.41 & 84.94±0.74 \\
& DGI & 93.09±0.08 & 92.19±0.34 & 89.19±0.67 & 88.03±0.28 & 92.79±0.22 & \underline{90.99±0.25} & \underline{89.36±0.12} \\
& COSTA & 92.67±0.16 & 91.18±0.31 & 86.90±0.64 & 85.39±0.47 & 92.41±0.20 & 90.07±0.57 & 87.77±0.16 \\ \cline{2-9}
& GraphMAE & 93.27±0.31 & \underline{92.93±0.52} & \underline{90.18±0.38} & \underline{88.22±0.76} & 91.57±0.36 & 87.59±2.09 & 82.74±1.13 \\
& MaskGAE & \underline{93.42±0.26} & 91.85±0.25 & 86.21±0.29 & 84.15±0.79 & \underline{93.06±0.24} & 90.57±0.38 & 87.80±0.18 \\ \cline{2-9}
& \textbf{\ourmethod(Ours)} & \textbf{94.49±0.17} & \textbf{93.27±0.17} & \textbf{91.19±0.21} & \textbf{91.03±0.25} & \textbf{93.74±0.31} & \textbf{92.01±0.14} & \textbf{90.91±0.23} \\ 
\midrule
\multirow{6}{*}{Computers}
& GRACE & 89.35±0.24 & 88.21±0.27 & 84.79±0.41 & 84.20±0.18 & 88.75±0.13 & 83.98±0.14 & 81.31±0.48 \\
& DGI & 89.60±0.22 & 88.29±0.06 & 84.24±0.22 & 82.70±0.42 & 88.95±0.10 & 86.53±0.18 & \underline{83.93±0.43} \\
& COSTA & 89.08±0.17 & 87.43±0.29 & 82.33±0.19 & 80.22±0.28 & 88.44±0.22 & 86.20±0.31 & 83.34±0.30 \\ \cline{2-9}
& GraphMAE & \underline{90.45±0.16} & \underline{89.57±0.24} & \underline{84.90±0.43} & \underline{84.42±0.21} & 89.56±0.24 & \underline{87.17±1.28} & 83.46±0.46 \\
& MaskGAE & 90.01±0.10 & 87.54±0.28 & 80.07±0.23 & 76.57±0.29 & \underline{89.57±0.08} & 85.59±0.24 & 81.96±0.39 \\ \cline{2-9}
& \textbf{\ourmethod(Ours)} & \textbf{91.37±0.09} & \textbf{89.87±0.17} & \textbf{87.72±0.24} & \textbf{87.29±0.21} & \textbf{90.26±0.24} & \textbf{87.74±0.18} & \textbf{85.73±0.33} \\
\bottomrule
\end{tabular}
}
\vspace{-3mm}
\label{tab:main}
\end{table*}

In this section, we evaluate the \ourmethod with extensive experiments on different datasets and noisy feature scenarios. Specifically, we aim to answer the following research questions:

\noindent\textbf{RQ1:} How {effective} is \ourmethod in various noisy feature scenarios?

\noindent\textbf{RQ2:} Can \ourmethod accurately estimate the noise intensity of nodes?

\noindent\textbf{RQ3:} How do pivotal designs impact the performance of \ourmethod?

\subsection{Experimental Setup}

\noindent\textbf{Datasets.} We use node classification as the downstream task to evaluate the informativeness of the learned representations. Our experiments are conducted on 5 benchmark datasets, including 3 citation graph datasets (i.e., Cora, CiteSeer, and PubMed~\cite{sen2008collective}) and 2 co-purchase graph datasets (i.e., Amazon Computers and Amazon Photo~\cite{shchur2018pitfalls}). All the datasets are randomly divided into 10\%, 10\%, and 80\% for training, validation, and testing respectively, following the setting in ~\cite{zhu2021graph,zhu2021empirical}.

\noindent\textbf{Evaluation Protocol.} For each experiment, we follow the evaluation scheme as in~\cite{zhu2020deep,velivckovic2018deep,zhang2022costa}. Specifically, each model is first trained in an unsupervised manner on the full graph with node features. Then, the learned representations are used to train and test an $l2$ regularized logistic regression classifier from SciKit-Learn~\cite{pedregosa2011scikit}. We conduct 5 runs for each model and report the mean classification accuracy along with its standard deviation as the metric.

\noindent\textbf{Feature Noise Injection.} Since all the datasets do not inherently contain feature noise, we manually injected two types of noise into them. Firstly, we randomly select $\alpha \in (0,1]$ fraction of nodes for perturbation. Then, we sample noise from either standard Gaussian or uniform distribution, and add them to the features of selected points with a coefficient of $\beta$ (i.e., noise level). 

\noindent\textbf{Baselines.} We selected representative UGRL methods as the baselines for comparison. Our baselines include 3 types of methods: 1) autoencoder-based UGRL methods, including GAE and VGAE~\cite{kipf2016variational}; 2) Contrastive UGRL methods, including DGI~\cite{velivckovic2018deep}, GRACE~\cite{zhu2020deep}, MVGRL~\cite{hassani2020contrastive}, and COSTA~\cite{zhang2022costa}; 3) Masked graph autoencoder-based UGRL methods, including GraphMAE~\cite{hou2022graphmae} and MaskGAE~\cite{,li2023s}.

\noindent\textbf{Implementation Details.} The experiments are conducted on a machine using Intel(R) Xeon(R) Platinum 8255C CPU@2.50GHz and a single RTX 4000 GPU with 16 GB GPU memory. The operating system of the machine is Ubuntu 20.04. We use Python 3.8, Pytorch 1.11.0, and CUDA 11.3 for the implementation of \ourmethod. Our code is available at~\url{https://github.com/Shiy-Li/MQE}.

\subsection{Performance Comparison (RQ1)}

\begin{figure} [tbp]
	\centering
	\subfloat[Normal Noise\label{subfig:proportion_normal}]{
            \includegraphics[height=0.375\columnwidth]{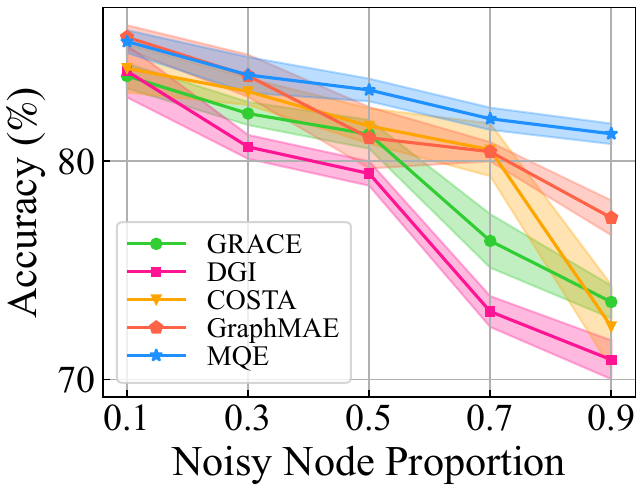}}\hfill
        \subfloat[Uniform Noise\label{subfig:proportion_uniform}]{
            \includegraphics[height=0.375\columnwidth]{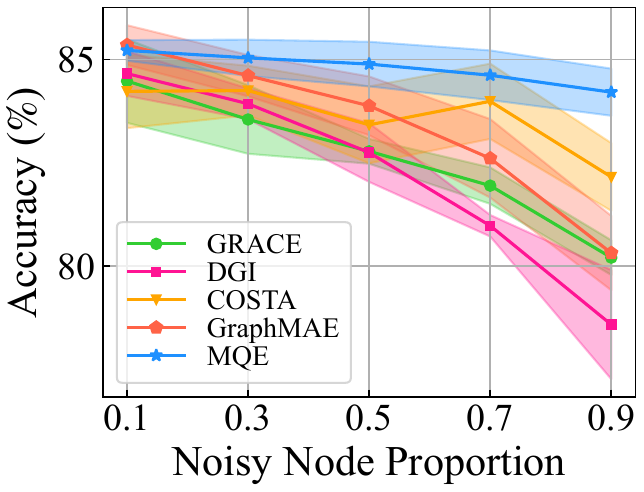}}\hfill

        \vspace{-2mm}
        \caption{Performance comparison in the scenarios with different noisy node fractions $\alpha$ on Cora dataset.}
        \vspace{-4mm}
	\label{fig:proportion}
\end{figure}

\noindent\textbf{Performance under different noise types and levels.} In this experiment, we established two distinct noise settings based on edge sparsity. Specifically, we fix fraction $\alpha=0.5$, vary $\beta=\{0.3, 0.5, 0.8\}$ (citation network), and vary $\beta=\{1, 5, 10\}$ (co-purchase network) for normal and uniform distribution to simulate different scenarios with noisy features. We also consider the scenario with clean data. On the three larger datasets, we only compare \ourmethod with competitive UGRL methods that have scalability, due to memory limitations. The experimental results are exhibited in Table~\ref{tab:main}, and from where we have the following observations. 1) \ourmethod generally outperforms the baselines in both clean and noisy feature scenarios. This superior performance demonstrates the effectiveness of multi-hop propagation and quality estimation in addressing the noisy feature issue. 2) Although we model the propagated features with a Gaussian distribution rather than a uniform one, \ourmethod still shows competitive performance in scenarios with uniform noise. This observation underscores the universality of \ourmethod, emphasizing its capability to learn informative representations from graph data with diverse feature noises. 3) As the noise level increases, the performance gap between \ourmethod and the baselines widens, which highlights the capability of \ourmethod in handling extremely noisy data. 

\noindent\textbf{Performance under different noisy node fraction $\alpha$.} To further verify the generalization ability of \ourmethod, we conduct this experiment by varying $\alpha$ from 0.1 to 0.9 while keeping $\beta=0.5$ constant. From the results in Fig.~\ref{fig:proportion}, we can find the significant performance advantage of \ourmethod when $\alpha$ is large. In contrast, the performance of some baselines (e.g., DGI) sharply declines as more nodes become noisy. \looseness-2

\subsection{Quality Estimation Effectiveness (RQ2)}\label{sec:qual_exp}

In this subsection, we conduct qualitative experiments to gain insights into the feature quality estimated by \ourmethod. We simulate the noisy feature scenario by setting $\alpha=0.5$ and $\beta=0.5$. 

\begin{figure} [tbp]
	\centering
	\subfloat[Cora\label{subfig:Cora_Info_Estima}]{
            \includegraphics[height=0.37\columnwidth]{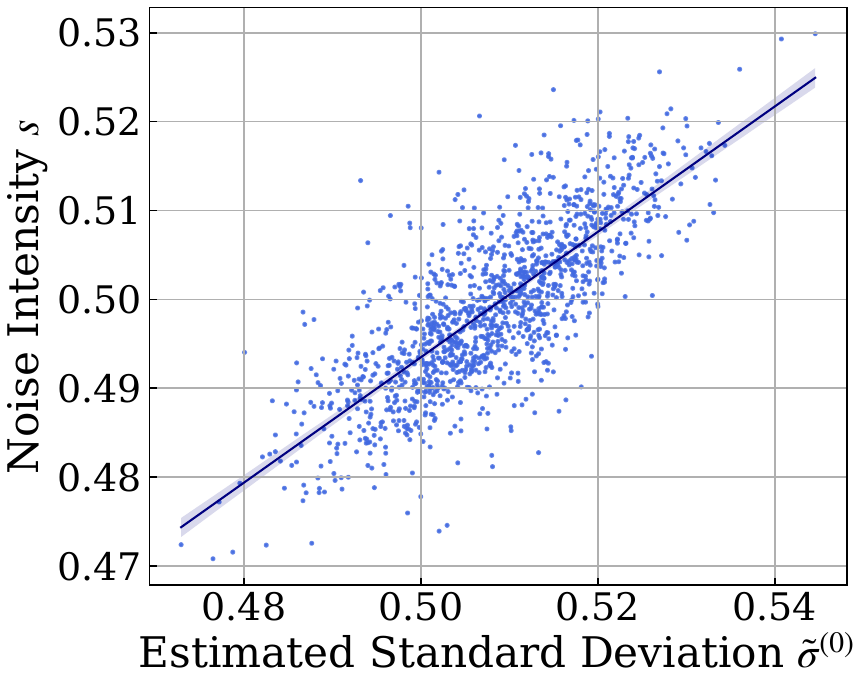}}\hfill
        \subfloat[CiteSeer\label{subfig:CiteSeer_Info_Estim}]{
            \includegraphics[height=0.37\columnwidth]{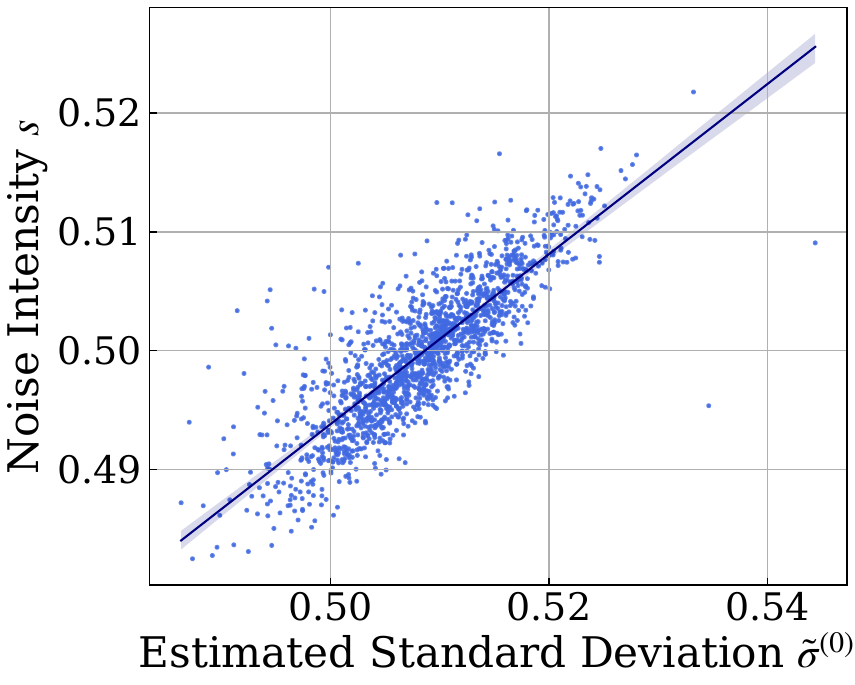}}\hfill
        \vspace{-2mm}
        \caption{The correlation between estimated standard deviation and feature noise intensity.}
        \vspace{-5mm}
	\label{fig:Info_Estima}
\end{figure}

\noindent\textbf{Intensity Estimation for Raw Features.} In Sec.~\ref{sec:pfqe}, we mention that the estimated standard deviation $\tilde{\sigma}_i^{(0)}$ at the $0$-th step propagated features (i.e., the raw features) can be an ideal indicator for noise intensity $s_i$. To verify the estimation effectiveness, we visualize the correlation between the noisy intensity (defined in Sec.~\ref{sec:pdef} and the estimated standard deviation in Fig.~\ref{fig:Info_Estima}. Note that we only visualize nodes with $s_i > 0$ and disregard those with clean features. The figure reveals a strong correlation between the ground-truth noisy intensity and the estimated $\sigma$, indicating that \ourmethod can effectively approximate the degree of feature noise for each node. This capability ensures that our method provides an in-depth understanding of the distribution of noisy features, thereby potentially aiding in data engineering and cleaning processes. 

\begin{figure} [tbp]
	\centering
	\subfloat[Step 0\label{subfig:Attack_Detect_MP_0}]{
        \includegraphics[height=0.28\columnwidth]{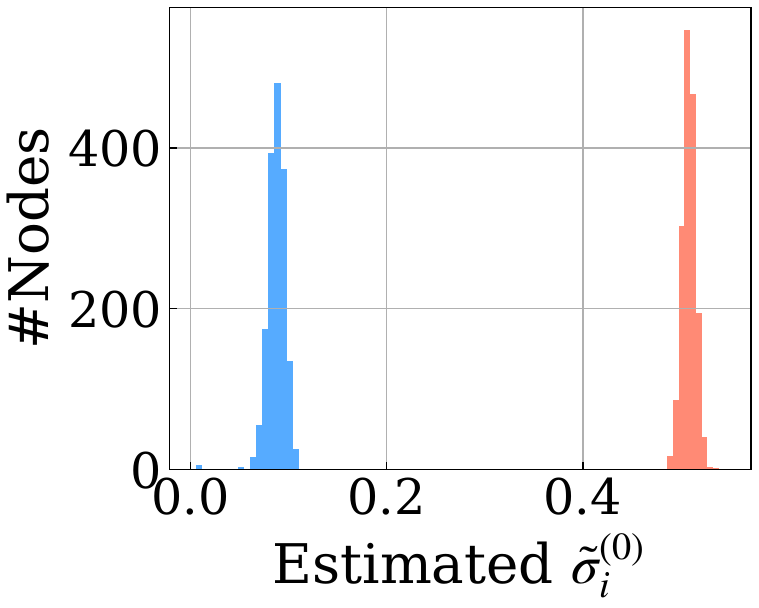}}\hfill
        \subfloat[Step 2\label{subfig:Attack_Detect_MP_1}]{
        \includegraphics[height=0.28\columnwidth]{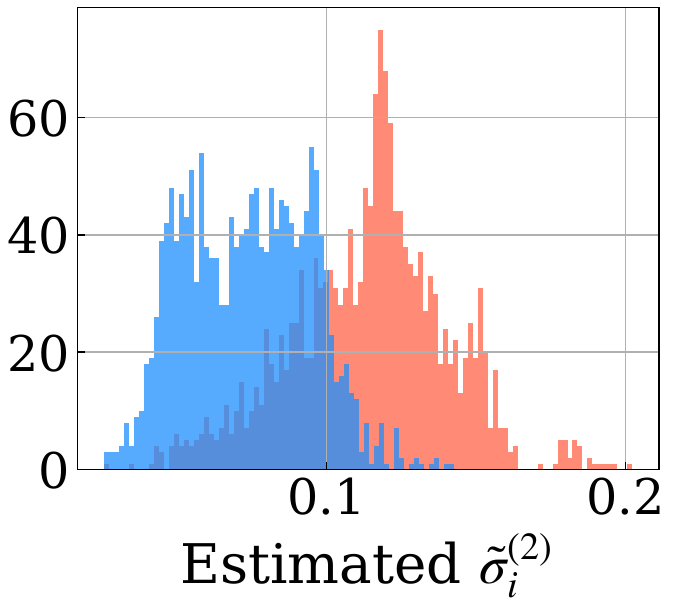}}\hfill
	\subfloat[Step 4\label{subfig:Attack_Detect_MP_2}]{
        \includegraphics[height=0.28\columnwidth]{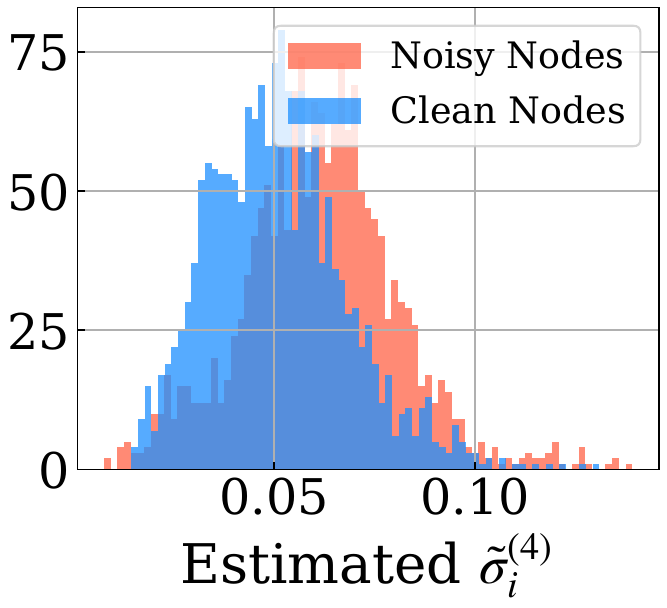}}
        
        \vspace{-2mm}
        \caption{Visualization of the distribution of estimated standard deviation at different propagation steps.}
        \vspace{-2mm}
	\label{fig:Attack_Detect}
\end{figure}

\noindent\textbf{Visualization of the Quality of Propagated Features.} By examining $\tilde{\sigma}_i^{(l)}$ estimated by \ourmethod at different propagation steps, we can further analyze the impact of propagation on noisy features. As depicted in Fig.~\ref{fig:Attack_Detect}, the estimated quality for the original features effectively distinguishes between clean nodes and noisy nodes. 
As the propagation process progresses, the estimated noise intensity of the noise nodes decreases and the noise intensity of some of the clean nodes increases, further confirming our observation and analysis in Sec.~\ref{MOTI}.

\subsection{Ablation Study (RQ3)}
To examine the contribution of each component and key design in \ourmethod, we conducted experiments on several variants of \ourmethod, and the results are shown in Table~\ref{Ablation}. Specifically, ``\ourmethod w/o aug'' represents the variant without graph structure augmentation; ``\ourmethod w/o mh'' is the variant that only estimates the quality at the last hop, and ``\ourmethod w/o reg'' removes the regularization loss term in Eq.~\eqref{eq.4-7}. We conduct experiments in both clean and noisy ($\alpha=0.5$, $\beta=0.8$, normal noise) feature scenarios.

From the results in Table~\ref{Ablation}, we have the following findings. 1) The performance of ``\ourmethod w/o mh'' is generally lower than \ourmethod, indicating the significance of considering the propagated information at multiple propagation steps. 2) Without the regularization term, the performance of \ourmethod can drastically degrade, which illustrates that the regularization term is critical in preventing trivial solution issues. 3) In most cases, augmenting the graph structure leads to improvement, particularly in scenarios with noisy features. 4) The performance degradation of the three variants of \ourmethod is generally higher in noisy scenarios than in clean scenarios, further illustrating the importance of these key designs in addressing the noisy feature problem.

\begin{table}
\centering
\caption{Performance of \ourmethod and its variants.}
\vspace{-2mm}
\begin{tabular}{p{1.4cm}|p{1.9cm}|p{1.7cm}<{\centering}|p{1.7cm}<{\centering}} 
\toprule
Dataset & Variant & \multicolumn{1}{c|}{Clean} & \multicolumn{1}{c}{Noisy} \\ 
\midrule
\multirow{4}{*}{Cora} & \ourmethod w/o aug & 84.61±0.33 & \underline{80.38±0.36} \\
 & \ourmethod w/o mh & \underline{85.38±0.46} & 79.53±0.64 \\
 & \ourmethod w/o reg & 81.96±1.12 & 30.89±0.51 \\ \cline{2-4}
 & {\ourmethod} & \textbf{85.97±0.42} & \textbf{82.16±0.55} \\ 
\midrule
\multirow{4}{*}{CiteSeer} & \ourmethod w/o aug & \underline{74.32±0.61} & 63.25±0.71 \\
 & \ourmethod w/o mh & 73.03±0.65 & \underline{64.42±0.88} \\
 & \ourmethod w/o reg & 53.28±4.98 & 21.54±1.01 \\\cline{2-4}
 & {\ourmethod} & \textbf{74.64±0.53} & \textbf{66.37±1.27} \\ 
\midrule
\multirow{4}{*}{PubMed} & \ourmethod w/o aug & \textbf{88.00±0.28} & \underline{76.48±0.31} \\
 & \ourmethod w/o mh & 83.93±0.36 & 76.13±0.31 \\
 & \ourmethod w/o reg & 85.84±0.35 & 51.63±0.91 \\\cline{2-4}
 & {\ourmethod} & \underline{87.73±0.19} & \textbf{80.37±0.38} \\
\bottomrule
\end{tabular}
\label{Ablation}
\end{table}

\section{Related Work}
Unsupervised graph representation learning (UGRL) aims to learn node-level representations from graphs by means of label-free cost, and is widely used in various real-world scenarios~\cite{ge2020graph, wang2021multi, zheng2022rethinking,liuyue_ELCRec}. Traditional methods such as deepwalk~\cite{perozzi2014deepwalk}, node2vec~\cite{node2vec_grover2016node2vec} generate low-dimensional embedding of nodes based on graph topology and random walk strategy. The recent popular contrastive UGRL methods are inspired by contrastive learning and follow the principle of maximizing mutual information to optimally learn node embeddings~\cite{velivckovic2018deep, hassani2020contrastive, zhu2020deep}. Another line of methods termed generative UGRL aims to learn node representations by recovering the missing parts of the input data~\cite{devlin2018bert,he2022masked,xie2022simmim,kipf2016variational}. For example, MGAE~\cite{tan2022mgae} and GraphMAE~\cite{hou2022graphmae} recover hidden structures and attributes by implementing masking strategies on graph structures and node attributes, allowing the model to obtain better node representations~\cite{li2023s}.

Unlike the aforementioned UGRL models, which are limited to perfect features, \ourmethod focuses on a broader spectrum of real-world scenarios with noisy features. At the methodology level, \ourmethod generates representations through the process of estimation rather than direct propagation via GNNs, which reduces the risk of noise diffusion during propagation.

\section{Conclusions}
In this paper, we undertake the first endeavor in unsupervised graph representation learning (URGL) on graph data with noisy features — a challenge that remains largely unexplored in real-world scenarios, despite its inevitability. Through empirical analysis, we uncover the advantages and drawbacks of message propagation in addressing the noisy feature issue, and recognize the importance of estimating the quality of propagated information. 
Based on these insights, we propose a new UGRL method, termed \ourmethod, which learns reliable node representations by estimating the quality of multi-hop propagated features with a conditional Gaussian model. Extensive experiments have demonstrated the effectiveness of \ourmethod in learning high-quality representations and approximating node-level noise intensity in various noisy feature scenarios.

\begin{acks}
This work was partially supported by the Specific Research Project of Guangxi for Research Bases and Talents GuiKe AD24010011, and the Innovation Project of Guangxi Graduate Education YCSW2024138.
\end{acks}

\bibliographystyle{ACM-Reference-Format}
\bibliography{sample-base}
\end{document}